\documentclass[10pt,twocolumn,letterpaper]{article}

\usepackage{cvpr}

\definecolor{cvprblue}{rgb}{0.21,0.49,0.74}
\usepackage[pagebackref,breaklinks,colorlinks,allcolors=cvprblue]{hyperref}
\usepackage{multirow} 

\usepackage{colortbl}  
\definecolor{Green}{RGB}{240, 255, 255}
\definecolor{background}{RGB}{237, 252, 214}
\definecolor{Gray}{RGB}{237, 252, 214}

\usepackage{algorithm}
\usepackage{algorithmic}

\def\paperID{}
\def\confName{CVPR}
\def\confYear{2026}

\title{Heuristic Self-Paced Learning for 
Domain Adaptive Semantic Segmentation under Adverse Conditions}

\newcommand{\whu}{1}%
\newcommand{\hubei}{2}%
\newcommand{\zhongguancun}{3}%
\newcommand{\whucs}{4}%
\newcommand{\hust}{5}%
\author{
Shiqin Wang\textsuperscript{\whu,\hubei,\whucs}\thanks{These authors contributed equally to this work.} \quad
Haoyang Chen\textsuperscript{\whu,\hubei,\zhongguancun,\whucs}\footnotemark[1] \quad
Huaizhou Huang\textsuperscript{\whu,\hubei,\whucs}\footnotemark[1] \quad
Yinkan He\textsuperscript{\whu,\hubei,\whucs} \quad
Dongfang Sun\textsuperscript{\whu,\hubei,\whucs} \\
Xiaoqing Chen\textsuperscript{\zhongguancun,\hust} \quad
Xingyu Liu\textsuperscript{\whu,\hubei,\whucs} \quad
Zheng Wang\textsuperscript{\whu,\hubei,\zhongguancun,\whucs}\thanks{Corresponding author.} \quad\quad
Kaiyan Zhao\textsuperscript{\whucs} \\
\normalsize{\textsuperscript{\whu}National Engineering Research Center for Multimedia Software, Institute of Artificial Intelligence, School of}\\
\normalsize{Computer Science, Wuhan University \quad \textsuperscript{\hubei}Hubei Key Laboratory of Multimedia and Network Communication Engineering}\\
\normalsize{\textsuperscript{\zhongguancun}Zhongguancun Academy, Beijing, China. 100094 \quad} \normalsize{\textsuperscript{\whucs}{School of Computer Science, Wuhan University, Wuhan, China}} \\
\normalsize{\textsuperscript{\hust}{School of Artificial Intelligence and Automation, Huazhong University of Science and Technology}} \\
}

\begin{document}

\maketitle
\begin{abstract}

The learning order of semantic classes significantly impacts unsupervised domain adaptation for semantic segmentation, especially under adverse weather conditions. Most existing curricula rely on handcrafted heuristics (e.g., fixed uncertainty metrics) and follow a static schedule, which fails to adapt to a model's evolving, high-dimensional training dynamics, leading to category bias. Inspired by Reinforcement Learning, we cast curriculum learning as a sequential decision problem and propose an \emph{autonomous class scheduler}. This scheduler consists of two components: (i) a high-dimensional state encoder that maps the model's training status into a latent space and distills key features indicative of progress, and (ii) a category-fair policy-gradient objective that ensures balanced improvement across classes. Coupled with mixed source–target supervision, the learned class rankings direct the network’s focus to the most informative classes at each stage, enabling more adaptive and dynamic learning. It is worth noting that our method achieves state-of-the-art performance on three widely used benchmarks (e.g., ACDC, Dark Zurich, and Nighttime Driving) and shows generalization ability in synthetic-to-real semantic segmentation.

\end{abstract}    
\section{Introduction}
\label{sec:intro}

\begin{figure}[!t]
\centering
\includegraphics[width=\columnwidth]{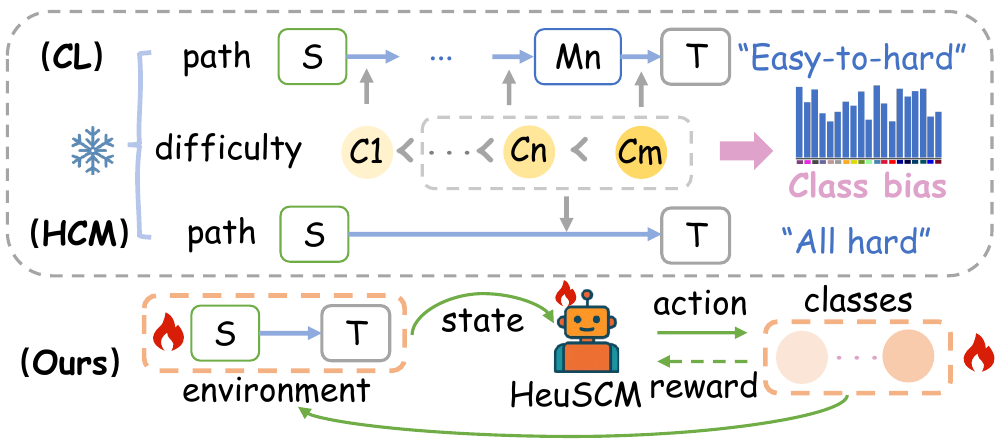}
\caption{From design curriculum to self-paced learning. Traditional Curriculum Learning (CL) and Hard Class Mining (HCM) both rely on fixed handcrafted priors and fixed learning paths. CL adopts a static, easy-to-hard curriculum, while HCM focuses solely on difficult classes. Both strategies induce class bias, manifesting as significant accuracy discrepancies across semantic classes. Differently, we adaptively mine the most informative classes via dynamically perceived models' evolving ($S$: Source domain, $T$: Target domain, $M_{n}$: the $n$-th intermediate domain, $Cn$: the $n$-th Semantic Class).
}
\label{problem}
\end{figure}

Semantic segmentation is one of the core technologies for autonomous driving systems to achieve robust environmental perception. However, existing models trained on controlled datasets, when deployed to the real world full of uncertainties, their performance will degrade significantly under real-world adverse weather conditions (such as heavy fog, nighttime, and heavy rain)~\cite{wang2023informative,wang2025parables,wang2024low,xu2021exploring,chen2025subjective}. Sensor data undergoes complex non-linear degradation under these conditions, directly leading to the failure of the model's ability to perceive safe-critical obstacles~\cite{zhong2022rainy,ma2022both,peng2021semantic}.

Unsupervised Domain Adaptation (UDA) is a mainstream paradigm to address domain shift issues and reduce reliance on target domain annotations. However, extreme weather poses an extraordinary compound challenge to UDA. This challenge consists of two intertwined problems: (1) Complex non-uniform domain shift: for example, fog concentration and rain intensity are continuously changing, resulting in feature shifts that are highly variable and non-linear; (2) Exacerbated class imbalance: In severe weather, the observability of inherently rare safety-critical semantic classes is further reduced, sharply amplifying the long-tail property of data distribution. 

Prevailing UDA methods attempt to tackle this compound challenge using separate, often decoupled, mechanisms. They typically employ style transfer~\cite{li2023vblc,zhengl2023compuda,li2024parsing,sakaridis2025condition} to mitigate the domain shift, while concurrently employing optimization strategies like Curriculum Learning (CL) and Hard Class Mining (HCS) to address the exacerbated class imbalance.
However, existing CL and HCS methods~\cite{gong2024coda} are generally limited by a common, fundamental paradigm flaw: they rely on a prior human definitions. Specifically, (1) the ``difficulty" of the curriculum is quantified by fixed artificially designed metrics (such as prediction uncertainty or confidence); (2) the ``path" of the curriculum is driven by artificially designed rules (such as ``from easy to hard" or ``all hard"). We believe that this ``prescriptive paradigm" is fundamentally unreasonable. The ``cognitive state" of a model during training is a high-dimensional, dynamic, and non-monotonic evolutionary process. Trying to statically plan this learning path with a fixed, one-dimensional, artificially defined scalar is suboptimal. The rigidity of this strategy makes it unable to adapt to the model's own ever-changing internal state, leading to unavoidable local optima when facing ``compound" problems like extreme weather, such as underfitting to noise or overfitting to the majority class, as shown in Figure~\ref{problem}.

Inspired by Reinforcement Learning (RL), we propose a paradigm shift from ``designing a curriculum" to ``learning a curriculum". We argue that the optimal learning trajectory should not be specified by human prior assumptions but should be autonomously discovered by the model based on its own learning state. In this paper, we groundbreakingly re-envision the training process of UDA as a ``sequential decision problem". We propose Heuristic Semantic Class Mining (HeuSCM), a framework that includes an autonomous scheduler (Agent). 
The efficacy of this method is built upon two core technical designs: \textit{i}. Autonomous State Perception: The decision-making of the Agent no longer relies on a single human-defined metric. We design a High-dimensional Semantic State Extraction (HSSE) network that enables it to comprehensively characterize the current learning progress of the model from high-dimensional state vectors.
 \textit{ii}. Dynamic Policy Optimization: The optimization objective of the Agent is not merely maximizing the summation of their individual value functions. We propose Categorical $\alpha$-Fairness for Policy Gradients (C$\alpha$PG), which optimizes the policy via our designed global fairness objective function for an equitable reward distribution across all semantic classes.

Our agent autonomously discovers a dynamic curriculum, adjusting in real-time according to the model’s state rather than following a pre-set path. This RL-inspired approach allows the model to adapt to complex, real-world scenarios, overcoming the limitations of traditional curriculum learning. Our main contributions are as follows:
\begin{itemize}
\item We are the first to redefine UDA curriculum learning from a ``human-defined heuristic" problem to an ``autonomously learned strategy" problem, enabling the model to dynamically adjust its learning path based on its evolving internal state.
\item We propose HeuSCM, a novel framework with the core being our HSSE and C$\alpha$PG design. HeuSCM realizes the faithful perception of high-dimensional semantic states and dynamic policy optimization that ensures categorical reward equity.
\item Extensive experiments on highly challenging extreme weather semantic segmentation benchmarks verify the effectiveness of our method, particularly achieving state-of-the-art performance 72.9 mIoU [\%] on the ACDC test. 
Moreover, our Heuristic Class Sampling Policy (HCSP) exhibits superior generalization capability on the synthetic-to-real segmentation benchmark.

\end{itemize}

\section{Related Work}
\label{sec:realted work}

\subsection{Unsupervised Domain Adaptation Semantic Segmentation Under Adverse Weather}
To bridge the large domain gap between the source domain (clear weather) and the target domain (adverse weather), some researchers generated the intermediate mixed domain via cross-domain mixed sampling and performed the domain adaptation from the source domain to the mixed domain~\cite{bruggemann2022refign,liu2024domain}. Other research efforts introduce style transfer~\cite{li2023vblc,zhengl2023compuda,li2024parsing,sakaridis2025condition} or image generation~\cite{shen2025w} networks to mitigate discrepancies in visual appearance, or alternatively focus on reducing domain-specific discrepancies at the feature level~\cite{lee2025frest}. 
Later, Bruggemann \textit{et al.}~\cite{bruggemann2023contrastive} designed Contrastive Model Adaptation (CMA) to learn domain-invariant features via aligning the features of target paired image pairs. However, these methods uniformly adapt source knowledge to adverse target weather, neglecting the inherent variations among challenging scenarios and causing the model to generate hallucinations (erroneous class predictions). To address this, Gong~\textit{et al.}~\cite{gong2024coda} introduced intermediate domains and performed the first-easy-then-hard domain adaptation (CoDA). 
Conversely, Chen~\textit{et al.}~\cite{chen2023amsc} focuses on learning hard classes that are visually similar within the target domain (AMSC). Regardless of the specific mechanism, whether easy-to-hard curriculum learning or hard class mining, it is strongly validated that the learning sequence of different samples or special attention to specific classes significantly promotes the field.
However, the difficulty assessment in these methods relies on a single, manually designed metric, and the learning path is fixed, often leading to insufficient model learning.

\subsection{Class Curriculum Learning} 
In cross-domain adaptation, methods concerning class curriculum learning primarily fall into Curriculum Learning (CL), sequencing learning from easy to hard, and Hard Class Mining (HCM), focusing on hard sample learning. These methods involve two main steps: difficulty assessment and curriculum scheduling. Regarding difficulty criteria, one line of research measures class difficulty based on target domain predictions, such as class frequency~\cite{bo2021hardness,liu2022hardboost,wang2023informative}, prediction uncertainty~\cite{wang2024curriculum}, or confidence~\cite{zhu2025hard}. Another line further incorporates domain discrepancy~\cite{wang2023informative} or feature similarity~\cite{wang2023informative1}. 
Regarding the learning scheduling, mainstream strategies can be categorized into three types. 
The first assigns selected classes higher sampling probabilities when performing cross-domain mixing sampling~\cite{wang2023informative,zhu2025hard}. The second assigns them higher loss weights to intensify the model's focus~\cite{zhang2023cross,liu2021bapa,wang2024curriculum}. The third directly utilizes these identified hard classes to retrain the model~\cite{bo2021hardness,liu2022hardboost}. 

Despite significant progress, the difficulty assessment criteria are manually fixed based on prior knowledge, and the class curriculum learning remains static. This prevents the model from adaptively learning the most informative classes according to its learning state, consequently resulting in insufficient learning of semantic classes.

\subsection{Reinforcement Learning in Unsupervised Domain Adaptation}
Reinforcement Learning (RL) has demonstrated its effectiveness in learning complex policies by interacting with the environment.
Some researchers~\cite{zhang2021adversarial,dong2020cscl,usmani2023reinforced,judge2025reinforcement} are exploring RL paradigms for unsupervised domain adaptation.
Zhang~\textit{et al.}~\cite{zhang2021adversarial} selected the most relevant cross-domain features via RL, and then applied adversarial learning to minimize the domain shift. This method focused on image classification.
In parallel, other studies~\cite{dong2020cscl,usmani2023reinforced,judge2025reinforcement} have explored RL paradigms for Unsupervised Domain Adaptation in Semantic Segmentation. Dong~\textit{et al.}~\cite{dong2020cscl} maximized the transfer gain under reinforcement learning manner. Usmani~\textit{et al.}~\cite{usmani2023reinforced} utilized RL techniques to realize cross-domain feature-level alignment. Judge~\textit{et al.}~\cite{judge2025reinforcement} introduced RL for 2D + time echocardiography segmentation.
Despite these efforts, most methods focus on feature alignment, with no attention to class sampling. 
\begin{figure*}[t]
  \centering
   \includegraphics[width=1.0\linewidth]{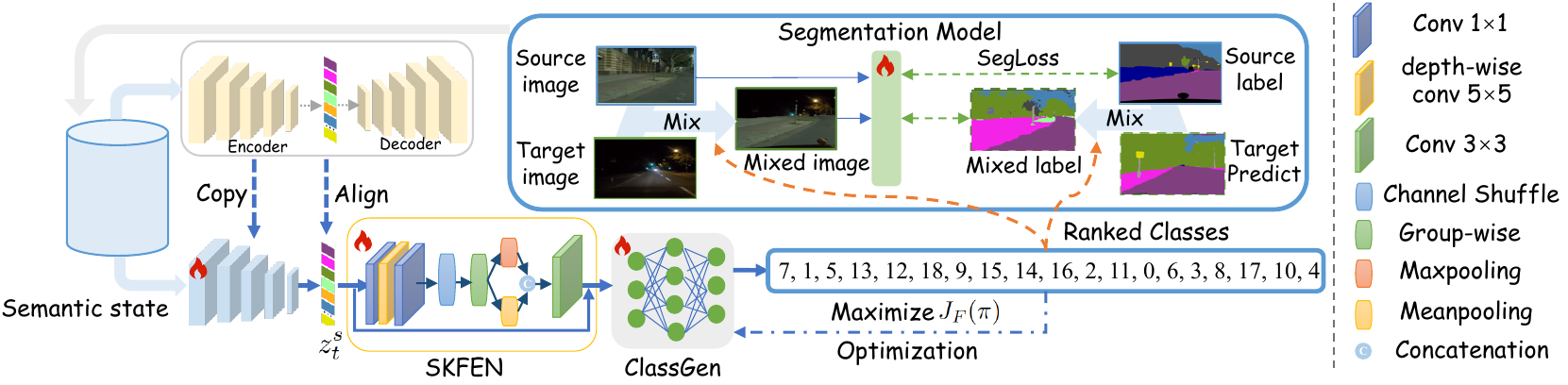}
   \caption{The framework of our designed Heuristic Semantic Class Mining (HeuSCM). First, a Gaussian Mixture VAE (GM-VAE) encodes high-dimensional semantic states into latent features $z_t^s$. Our SKFEN processes $z_t^s$ to distill key features reflecting the model's learning status. Conditioned on these features, ClassGen outputs ranked classes (sorted in descending order of informational value). These rankings guide the generation of mixed image and mixed label pairs, which optimize the segmentation model via SegLoss. Concurrently, we maximize the objective $J_F(\pi)$ to jointly optimize the copied GM-VAE encoder, SKFEN, and ClassGen.}
   \label{fig:framework}
\end{figure*}

\section{Method}

In this paper, we propose a Heuristic Semantic Class Mining (HeuSCM) self-paced curriculum framework, conceptually inspired by reinforcement learning, that dynamically perceives the learning progress of the semantic segmentation model from a high-dimensional, evolving state space.
Instead of relying on fixed, predefined rules, HeuSCM learns a policy-based class scheduler that autonomously adjusts the learning strategy based on a reward signal, rather than a static curriculum.
As illustrated in Figure~\ref{fig:framework}, we first obtain the learning status of the current semantic segmentation model from the high-dimensional segmentation state (Sec .~\ref{method1}). Then we continuously optimize ClassGen to generate ranked semantic classes (sorted in descending order of informativeness, Sec .~\ref{method2}), which guide the cross-domain mixed sampling and in turn update the segmentation model via the SegLoss.

\noindent\textbf{Segmentation Model:} 
UDA-SS under adverse weather aims to adapt the semantic segmentation knowledge from the source domain $D_{S}=  \left\{(x^{i}_{s},y^{i}_{s})\right\}^{N_{s}}_{i=1}$ to the target domain $D_{T}= \left\{(x^{j}_{t})\right\}^{N_{t}}_{j=1}$.
To bridge the large domain gap, cross-domain mixed sampling between the $D_{S}$ and $D_{T}$ is performed to generate the mixed domain $D_{M}=  \left\{(\mathcal{X}^{t}_{Tmix},\mathcal{Y}^{t}_{Rmix})\right\}^{N_{m}}_{t=1}$ as the bridge, and then $D_{S} \rightarrow D_{M}$ domain adaptation is performed. $N_{s}$, $N_{t}$, and $N_{m}$ denote the source, target domain, and mixed domain sample sizes.

The $t$-th mixed image ($\mathcal{X}^{t}_{Tmix}$) and label ($\mathcal{Y}^{t}_{Rmix}$) is generated as follows:
\begin{equation}
\left\{\begin{array}{l}
\begin{aligned}
& \mathcal{X}^{t}_{Tmix} = \mathcal{H}\odot x^{i}_{s} + \left(1-\mathcal{H}\right)\odot x^{j}_{t}, \\
& \mathcal{Y}^{t}_{Rmix} = \mathcal{H}\odot y^{i}_{s} + \left(1-\mathcal{H}\right)\odot g_\theta(x^{j}_{t})^{r}, \\
& \mathcal{H}_{ijk} = \mathbb{I}(L_S(m, n) = k) \cdot \mathbb{I}(k \in \mathcal{C}_{\text{low}}), \\
& \mathcal{C}_{\text{low}} = \{ \mathcal{R}^{i}_S(k) \mid \lfloor N^{i}_S / 2 \rfloor + 1 \le k \le N^{i}_S \}, \\
& \mathcal{R}^{i}_S = (c_k \in \mathcal{R} \mid c_k \in C^{i}_S)), \\
\end{aligned}
\end{array}\right.
\label{mixup}
\end{equation}
where $g_\theta$, $x^{i}_{s}$, $x^{j}_{t}$, $y^{i}_{s}$ denote the segmentation model, the $i$-th source image, $j$-th target image, and $i$-th source label, respectively. $\mathcal{H} \in \{0, 1\}^{h \times w \times c}$, $\mathbb{I}$ are the binary mask and an indicator function, respectively. 
$(m,n)$ denotes the spatial coordinates of the pixel, and $k$ is the semantic class index. $C^{i}_S$ and $N^{i}_S$ denote the semantic classes and classes number of the $i$-th source label. $\mathcal{R}$ is the ``Ranked Classes". The core mechanism of HeuSCM is that this list $\mathcal{R}$ is not fixed; it is the action output by our autonomous agent, dynamically generated based on the model's current state.

After that, the segmentation model ($g_\theta$) is optimized via the following training loss function (\textit{i.e.,} SegLoss):
\begin{equation}
\resizebox{0.9\columnwidth}{!}{$
\begin{aligned}
\mathcal{L}_{seg}=& \lambda_{1}\mathcal{L}_{CE}\left(g_\theta(x_{s}),y_{s}\right) + \lambda_{2}\mathcal{L}_{CE}\left(g_\theta(\mathcal{X}_{Tmix}),\mathcal{Y}_{Rmix}\right), \\
\end{aligned}
$}
\label{segloss}
\end{equation}
where $\mathcal{L}_{CE}$ is cross-entropy loss. $\lambda_{1}$, $\lambda_{2}$ are the coefficient.

\subsection{High-dimensional Semantic State Extraction}
\label{method1}
\subsubsection{Low-dimensional State Representation Learning}
\label{low-dimensional}  
The state space reflecting the semantic segmentation model's learning status is high-dimensional and redundant, comprising coupled features and complex interactions that jointly capture the subtle nuances of the domain adaptation process. Direct learning within such a space presents significant challenges for policy network optimization. 
To tackle this, our goal is to construct a low-dimensional latent state representation that effectively captures the intricate relationships among these high-dimensional features. 

To model the multi-modal nature of the semantic segmentation model's learning, we employ a Gaussian Mixture Variational Autoencoder (GM-VAE) to encode the high-dimensional learning features into a compact latent space, where each dimension represents a distinct aspect of the model's domain adaptation status. During training, given a high-dimensional state $\mathbf{s} \in \mathcal{S}$, the GM-VAE encoder ($\text{Enc}_{\psi}$) learns to infer the posterior over both the discrete component, \textit{i.e.}, categorical distribution $q_{\psi}(c|\mathbf{s}) = \text{Cat}(c; \text{Enc}_{\psi}(\mathbf{s}))$ (representing the learning mode), and the continuous latent state posterior $q_{\psi}(\mathbf{z}|\mathbf{s}, c) = \mathcal{N}(\mathbf{z}; \mu_{\psi}(\mathbf{s}, c), \Sigma_{\psi}(\mathbf{s}, c))$. The model is trained by maximizing the following variational lower bound:
\begin{equation}
\resizebox{0.9\columnwidth}{!}{$\displaystyle
\begin{split}
\mathcal{L}_{\text{GM-VAE}}(\mathbf{s}) & = \mathbb{E}_{q_{\psi}(c|\mathbf{s})} \left[ \mathbb{E}_{q_{\psi}(\mathbf{z}|\mathbf{s}, c)} \left[ \log p_{\theta}(\mathbf{s}|\mathbf{z}) \right] \right. \\
& \quad \left. - \text{KL}\left(q_{\psi}(\mathbf{z}|\mathbf{s}, c) \| p(\mathbf{z}|c)\right) \right] - \text{KL}\left(q_{\psi}(c|\mathbf{s}) \| p(c)\right),
\end{split}
$}
\label{GM-VAE}
\end{equation}
where $p_{\theta}$ is the probabilistic decoder which reconstructs the original state $\mathbf{s}$ from the latent variable $\mathbf{z}$. $q_{\psi}$ represents the approximate posteriors and $\text{KL}(\cdot \| \cdot)$ denotes the Kullback–Leibler divergence. $p(\mathbf{z}|c)$ and $p(c)$ are the priors.

After unsupervised pre-training, we freeze the GM-VAE decoder and retain the encoder to perform the mapping from the high-dimensional state space to the low-dimensional latent space. To accommodate task-specific requirements, we jointly fine-tune the encoder together with the policy network optimization. As the policy adjusts based on the reward signal, the encoder is co-updated to align the latent learning state with the reward-relevant semantic class outputs. At the same time, we regularize the encoder so that the learned latent representation still preserves the structure of the original state manifold, via the  reconstruction loss as follows:
\begin{equation}
\mathcal{L}_{\text{recon}}
= \mathbb{E}_{\mathbf{s}_t \sim \mathcal{D}_{\text{state}}}
\left[ \left\| \mathbf{s}_t - p_{\theta}\big(\text{Enc}_{\psi}(\mathbf{s}_t)\big) \right\|^2 \right],
\label{recon}
\end{equation}
where $\mathbf{s}_t$ denotes the high-dimensional segmentation state at step $t$, $\text{Enc}_{\psi}$ is the encoder of the copied GM-VAE, and $p_{\theta}$ is the frozen decoder.

\subsubsection{Semantic Key Feature Extraction Network}
\label{SKFEN} 

To address the large redundant information within the low-dimensional latent space and distill the key features that truly reflect the model’s learning status, we propose the Semantic Key Feature Extraction Network (SKFEN). Inspired by traditional spatial pooling and grouped convolution, SKFEN does not operate on spatial compression; rather, we introduce a novel feature refinement mechanism designed innovatively from the perspective of reducing feature redundancy among different channels. 

Our SKFEN contains two main phases: (1) Initial Transformation and Grouping: the low-dimensional state $z_t^s$ first undergoes initial feature fusion and interaction modeling. Subsequently, its channel dimension is expanded and permuted to prepare for grouped processing as follows: 
\begin{equation}
\resizebox{0.9\columnwidth}{!}{$
\begin{aligned}
f_{\text{group}} = \text{Group}(&\text{Shuffle}( \text{Conv}_{1 \times 1}^{\text{expand}}(\text{Conv}_{5 \times 5}^{\text{ds}} (\text{Conv}_{1 \times 1}^{\text{fuse}}(z_t^s))) ) ) = \bigcup_{g=1}^{G} f_g,
 \end{aligned}
 $}
 \label{eq:skfen_grouping}
\end{equation}
where $z_t^s$ is the input low-dimensional latent state, $\text{Conv}_{1 \times 1}^{\text{fuse}}(\cdot)$ denotes an initial $1 \times 1$ convolution for feature fusion, $\text{Conv}_{5 \times 5}^{\text{ds}}(\cdot)$ represents a $5 \times 5$ depth-wise separable convolution for spatial interaction modeling, and $\text{Conv}_{1 \times 1}^{\text{expand}}(\cdot)$ is a $1 \times 1$ convolution that expands the channel dimension to $n$. $\text{Shuffle}(\cdot)$ is the channel shuffle operation, $\text{Group}(\cdot)$ splits the feature map into $G$ groups along the channel dimension, resulting in feature maps $f_g$ for each group $g \in \{1, \dots, G\}$.

(2) Group-wise Feature Aggregation: These grouped features $f_g$ are then processed in parallel to distill salient and statistical information. These two distinct representations are concatenated and fused by a final convolution to obtain the refined features. This entire aggregation and fusion process is formulated as:
\begin{equation}
\resizebox{0.77\columnwidth}{!}{$\displaystyle
\begin{split}
z_{\text{out}} = & \text{Conv}_{3 \times 3} \left( \text{Concat} \left( \bigcup_{g=1}^{G} \left[\max_{c=1}^{C_g} { f_g(c, \cdot, \cdot) }\right], \right. \right. \\
& \left. \left. \bigcup_{g=1}^{G} \left[\frac{1}{C_g} \sum_{c=1}^{C_g} f_g(c, \cdot, \cdot)\right] \right) \right) + z_t^s,
\end{split}
$}
\label{eq:skfen_aggregation}
\end{equation}
where $z_{\text{out}}$ is the final refined latent state, and $z_t^s$ is the original input state from Eq. \eqref{eq:skfen_grouping} used for the residual connection. $\text{Conv}_{3 \times 3}$ denotes the final fusion convolution. For each feature group $f_g$ (with $C_g = n/G$ channels) obtained from $f_{\text{group}}$, the term $\left[\max_{c=1}^{C_g} \{ f_g(c, \cdot, \cdot) \}\right]$ represents the channel-wise max pooling operation, which computes the maximum value across all channels $c$ for each spatial location. Similarly, $\left[\frac{1}{C_g} \sum_{c=1}^{C_g} f_g(c, \cdot, \cdot)\right]$ represents the channel-wise average pooling operation. The $\bigcup_{g=1}^{G}$ operator signifies concatenation along the channel dimension across all $G$ groups, and $\text{Concat}(\cdot)$ concatenates the resulting max-pooled and average-pooled feature maps.

\begin{algorithm}[t] 
\caption{Heuristic Class Mining.} 
\label{alg:Framwork} 
\begin{algorithmic}[1] 
\STATE \textbf{Initialize:} Segmentation Model $g_\theta$, GM-VAE encoder $\text{Enc}_{\psi}$ and decoder $p_{\theta}$, SKFEN, policy network (ClassGen) $\pi$, Replay Buffer $\mathcal{D}$, Value (Critic) Networks $V_{c, \xi}$ (for $c=1..C$)
\FOR{$t\gets$ 1 to $T$}
\STATE Observe high-dim state $\mathbf{s}$ from $g_\theta$
\STATE Encode low-dim state $z_t^s = Enc_{\psi}(\mathbf{s})$
\STATE Distill key features $z_{key}$  \qquad \qquad $\triangleright$ Eq. (\ref{eq:skfen_grouping}), (\ref{eq:skfen_aggregation})
\STATE Generate Ranked Classes $\mathcal{R}_t = \pi_{\theta}(z_{key})$
\STATE Generate mixed data $(\mathcal{X}_{Tmix}, \mathcal{Y}_{Rmix})$ $\triangleright$ Eq.(~\ref{mixup})
\STATE Update $g_\theta$ by minimizing $\mathcal{L}_{seg}$ \qquad $\triangleright$ Eq. (\ref{segloss})
\STATE Compute reward $\vec{r}_t = [r_1(t), ..., r_C(t)]$ $\triangleright$ Eq. (\ref{reward})
\STATE Observe next state $z_{key, t+1}$
\STATE Record transition $(z_{key, t}, \mathcal{R}_t, \vec{r}_t, z_{key, t+1})$ in $\mathcal{D}$
\STATE Record state $\mathbf{s}_t$ in $\mathcal{D}_{\text{state}}$
\ENDFOR \\
\STATE \quad \textbf{for} $t$ = 1 to $T_{\text{agent\_step}}$ \textbf{do}
\STATE \qquad Sample a batch of transitions from $\mathcal{D}$
\STATE \qquad Update $\pi_{\theta}$, SKFEN, $\text{Enc}_{\psi}$ via policy gradient $\nabla_{\theta} J_F(\pi_{\theta})$ $\triangleright$ Eq. (~\ref{policy_gradient})
\STATE \qquad Sample a batch of states $\mathbf{s}_t$ from $\mathcal{D}_{\text{state}}$
\STATE \qquad Update $\text{Enc}_{\psi}$ by minimizing $\mathcal{L}_{recon}$ \qquad $\triangleright$ Eq. (\ref{recon})
\STATE \quad \textbf{end for}
\STATE \textbf{end for}
\end{algorithmic}
\end{algorithm}

\begin{table*}
\caption{Comparison of the state-of-the-art in Cityscapes$\rightarrow$ACDC domain adaptation on the ACDC test set. The best results are presented in bold.}
\renewcommand\arraystretch{1.2}
\resizebox{\linewidth}{!}{
\begin{tabular}{cccccccccccccccccccccc}
\hline
Method                    & \rotatebox{90}{Backbone}  &\rotatebox{90}{road}          &\rotatebox{90}{sidewalk}       &\rotatebox{90}{building}      &\rotatebox{90}{wall}         &\rotatebox{90}{fence}         &\rotatebox{90}{pole}          &\rotatebox{90}{traffic light} &\rotatebox{90}{traffic sign}  &\rotatebox{90}{vegetation}     &\rotatebox{90}{terrain}       &\rotatebox{90}{sky}           &\rotatebox{90}{person}        &\rotatebox{90}{rider}          &\rotatebox{90}{car}            &\rotatebox{90}{truck}          &\rotatebox{90}{bus}           &\rotatebox{90}{train}          &\rotatebox{90}{motorcycle}     &\rotatebox{90}{bicycle}        & \textbf{mIoU} \\ \hline
DeepLab-v2~\cite{chen2017deeplab}             & DeepLab-v2   & 71.9          & 26.2           & 51.1          & 18.8          & 22.5           & 19.7          & 33.0          & 27.7          & 67.9          & 28.6         & 44.2          & 43.1         & 22.1         & 71.2         & 29.8         & 33.3         & 48.4          & 26.2         & 35.8         & 38.0          \\
Refign~\cite{bruggemann2022refign}             & DeepLab-v2   & 49.5           & 56.7           & 79.8          & 31.2          & 25.7           & 34.1          & 48.0          & 48.7          & 76.2          & 42.5         & 38.5          & 48.3          & 24.7         & 75.8         & 46.5         & 43.9         & 64.3          & 34.1         & 43.6         & 48.0          \\
CMA~\cite{bruggemann2023contrastive}             & DeepLab-v2   & 83.1            & 52.7            & 65.4           & 18.7          & 30.5            &   \textbf{44.5}        & 56.3           &  53.9         &  76.7         &  39.7        &  79.0         &   54.2        &  31.2        &  76.7        &  40.2        &  39.3        &  47.4        &  29.8        &  38.6        &  50.4         \\
CompUDA~\cite{zhengl2023compuda} & DeepLab-v2    & 52.4           & 54.5          & 75.6         & 30.6           & 26.8          & 35.6        & 44.7         & 47.8         & 74.5          & 40.5        & 39.1          & 45.1         & 20.6        & 76.3          & 47.2         & 40.5          & 64.9         & 36.2         & 36.2          &  47.0        \\ 
VBLC~\cite{li2023vblc}             & DeepLab-v2   & 49.6           & 39.3           & 79.4          & 35.8          & 29.5           & 42.6          & 57.2          & 57.5          & 69.1          & 42.7         & 39.8          & \textbf{54.5}          & 29.3         & 77.8         & 43.0         & 36.2         & 32.7          & 38.7         & \textbf{53.4}         & 47.8          \\

ATP~\cite{wang2024curriculum}   & DeepLab-v2    &  76.2          &   47.3        & 71.4          &  \textbf{42.7}         &  31.4          &  44.2         &  55.4         &  \textbf{62.0}        &  \textbf{89.0}         &  34.7        &  79.1         &  49.9         &  16.6        &  77.5        &  30.0        &  19.7       &  47.7         & \textbf{44.0}        &  39.4        &  50.5         \\

CISS~\cite{sakaridis2025condition} & DeepLab-v2    & 70.5            &  36.7         &  67.0         &  29.4         & 30.2           &  31.6         & 45.6          &  48.9        &  70.4        &  24.7        &  65.5         &  48.2         &  31.1        &  76.6        &  45.7        &  47.0       &  62.8         &  26.8      &  38.9       &  47.2         \\

\rowcolor{Green} HeuSCM (Ours)             & DeepLabv2   &  \textbf{91.4}         &  	\textbf{66.6}           &  	\textbf{84.3}        &   	40.6         &   \textbf{32.3}       &  	38.0         &  	\textbf{57.9}        &    54.8      &   	82.8      &   \textbf{50.4}     &  \textbf{94.2}         &   	53.8      &    	\textbf{33.0}    &    \textbf{80.1}    &  \textbf{50.8}       &   \textbf{52.4}       &    \textbf{72.2}      &  29.3      &  	50.7    &   \textbf{58.7}      \\

\hline
DAFormer~\cite{hoyer2022daformer}             & DAFormer   & 56.9          & 45.4           & 84.7          & 44.7          & 35.1           & 48.6          & 44.8          & 57.4          & 69.5          & 52.9         & 45.8          & 57.1         & 28.2         & 82.8         & 57.2         & 63.9         & 84.0          & 40.2         & 50.5         & 55.3          \\ 
Gaussian~\cite{yu2025contrastive} & DAFormer   &  62.8         &  51.6         &  83.0        &  34.7        &  35.0        &  52.1        &  30.1       &  56.4        &  73.0       &  55.9       &  60.9         &  62.7       &  33.8        &    80.2     &  59.5       &  58.5      &  81.8       &  47.5       &  52.3        & 56.4          \\ 
Refign~\cite{bruggemann2022refign}             & DAFormer   & 89.5           & 63.4          & 87.3          & 43.6          & 34.3           & 52.3          & 63.2          & 61.4          & 86.9          & 58.5         & 95.7          & 62.1          & 39.3         & 84.1         & 65.7        & 71.3         & 85.4         & 47.9         & 52.8         & 65.5          \\ 
VBLC~\cite{li2023vblc}             & DAFormer    & 89.2   &  59.8          & 85.9           & 44.0            &  37.2           &  53.5            &  \textbf{64.5}          &  63.2         &  72.4          & 56.3           & 84.1          &  65.5         & 37.7         &  85.1         &  60.1       & 71.8         &  85.2         &  47.7       &  56.3               &  64.2        \\
CoPT~\cite{mata2025copt}  & DAFormer    & 49.1   &  \textbf{70.3}          &   83.6         &   \textbf{59.4}          &   \textbf{42.4}          &   \textbf{58.5}           &  48.3          &  \textbf{67.2}          &   73.5         &   \textbf{60.7}         &  45.0         &   \textbf{69.3}        &  \textbf{45.2}        &   83.4        &   \textbf{76.3}      &   \textbf{74.5}       &   \textbf{88.2}       &  \textbf{54.4}       &   \textbf{61.4}              &  63.7        \\
Instance-Warp~\cite{zheng2025instance}  & DAFormer   &  83.0        & 53.2           &  85.5          &   47.4          &  38.3            &  46.0          & 51.4          &    57.8        &   73.9         &  56.2          &  82.1          &    61.3      &  32.3         &  \textbf{85.5}       &  69.0        &  68.9         &  82.5       & 46.7       &  52.0       &  61.7         \\
\rowcolor{Green} HeuSCM (Ours)             & DAFormer   &  \textbf{90.2}         &  64.4           &  \textbf{87.3}        &   43.5         &   34.9       &  51.9         &  63.6         &   61.6        &   \textbf{86.9}      &  59.1      &  \textbf{95.8}         &  62.1        &  39.2      & 84.3        &   65.4      &  71.3        & 85.4         &  48.3       &  52.2      &     \textbf{65.7}    \\
\hline

HRDA~\cite{hoyer2022hrda}             & HRDA   & 88.3           & 57.9           & 88.1          & 55.2          &  36.7          & 56.3          & 62.9          & 65.3          & 74.2          & 57.7         & 85.9          & 68.8        & 45.7         & 88.5          & 76.4        & 82.4         & 87.7         & 52.7          & 60.4        &  68.0         \\
Refign~\cite{bruggemann2022refign}             & HRDA    & 93.8           & \textbf{75.7}           &  90.0           & 57.9            &  43.3            & 55.6            & 67.4          & 68.2           & 88.2           & 61.8          & 96.1          & 67.5         & 50.8          & 88.8        & 75.2         & 83.4          & \textbf{89.6}          & 54.5        & 61.2         &  72.1         \\
CompUDA~\cite{zhengl2023compuda} & HRDA    &  92.7          & 71.5           & 89.5         & \textbf{61.6}           &  39.8         &  51.0    & \textbf{72.0}  & 67.2  & 82.8 & 58.7 & 92.9 & 67.0 & 46.4 & 89.3 & 75.3 & 81.2 & 88.7 &  56.3 & 62.4  &  71.1        \\ 
VBLC~\cite{li2023vblc}             & HRDA     & 90.2           &  63.9          & 87.8            &  44.1           & 42.3             & 54.1           & 67.0          &  65.5          & 74.4           &  58.9         &  85.9         &  66.4        & 43.8          & 87.5        & 72.0         &  83.9        &  84.2        &  48.5      &  57.1        &  67.2        \\
CoDA~\cite{gong2024coda} & HRDA     & 93.1  & 72.7   & \textbf{90.7}  & 57.3   & \textbf{47.4}   &  56.8  & 69.9    & 70.0     & 87.3    & 59.8   & 95.4   & \textbf{71.4}  & 47.6  & \textbf{90.3}  & \textbf{77.1}  & 83.8 & 89.1  & 54.7 & \textbf{64.1}  & 72.6        \\
ACSegFormer~\cite{liu2024domain} & HRDA   & \textbf{94.3}  &  75.3  & 90.2  & 57.9   & 42.1    & 55.7    & 71.4    & \textbf{71.6}     & 87.8    & 61.5   & 96.0   & 69.8  & 51.2  & 88.7  & 75.5  & 85.8  & 89.5  & 53.8  & 62.3  & 72.7        \\ 
CISS~\cite{sakaridis2025condition} & HRDA      &  92.0   & 69.6           &  89.2           &  57.3            &  40.5           &  55.8             &   67.1         &  67.3         &   75.3         &   59.7        & 86.4       &  70.0         &   47.5       &   88.9        &   73.1      &  77.5         &  87.0        &  55.6       &   61.7              &  69.6       \\
\rowcolor{Green} HeuSCM (Ours)             & HRDA   &   93.6
         &  74.6           &  90.5     &   59.9   &  42.7        &    \textbf{57.3}       &   71.1    &    70.3    &   \textbf{88.7}     &    \textbf{62.4}    &  \textbf{96.4}  &   70.9        &  \textbf{52.3}       &  88.7    &   74.9     &  \textbf{86.3}       &  87.0        &    \textbf{57.3}     &  61.0     &  \textbf{72.9}         \\

\hline

\end{tabular}
}
\label{t:acdctest}
\end{table*}

\begin{table*}
\caption{Comparison of the state-of-the-art in Cityscapes$\rightarrow$ACDC domain adaptation on the ACDC val set. The best results are presented in bold.}
\renewcommand\arraystretch{1.2}
\resizebox{\linewidth}{!}{
\begin{tabular}{ccccccccccccccccccccccc}
\hline
Method                    & \rotatebox{90}{Backbone}   &\rotatebox{90}{road}          &\rotatebox{90}{sidewalk}       &\rotatebox{90}{building}      &\rotatebox{90}{wall}         &\rotatebox{90}{fence}         &\rotatebox{90}{pole}          &\rotatebox{90}{traffic light} &\rotatebox{90}{traffic sign}  &\rotatebox{90}{vegetation}     &\rotatebox{90}{terrain}       &\rotatebox{90}{sky}           &\rotatebox{90}{person}        &\rotatebox{90}{rider}          &\rotatebox{90}{car}            &\rotatebox{90}{truck}          &\rotatebox{90}{bus}           &\rotatebox{90}{train}          &\rotatebox{90}{motorcycle}     &\rotatebox{90}{bicycle}        & \textbf{mIoU} \\ \hline
DeepLab-v2~\cite{chen2017deeplab}             & DeepLab-v2   & 69.5           & 15.3           & 52.4            & 11.0            & 12.0             & 25.4           & 48.2          & 39.6           & 68.3           & 24.2          & 69.0          & 40.1         & 15.2          & 64.7        & 14.2         & 19.8          & 28.8         & 21.8       & 17.2         & 34.6          \\
Refign~\cite{bruggemann2022refign}             & DeepLab-v2   &   91.5       &  67.5         &  \textbf{81.4}           &  43.8           &  32.5            &  43.0          & 66.4          &  50.1          &  \textbf{82.1}         & 35.0         &  93.3         &   47.7       &  27.4         &  78.7       &  52.4        &  51.8         &  \textbf{64.1}        &   13.9     &  \textbf{37.7}       &   55.8       \\
CMA~\cite{bruggemann2023contrastive}             & DeepLab-v2  & 83.9           & 51.3           & 60.1            & 25.3            & 28.2             & 45.6           & \textbf{69.1}          & 52.3           & 79.0           & 30.1          & 79.2          & 51.5         & 23.5          & 77.4        & 43.4         & 37.0          & 36.2         & 24.9       & 28.6         & 48.8        \\
VBLC~\cite{li2023vblc}             & DeepLab-v2     & 50.0           & 42.5           &  73.6           &  34.5           &  26.3            & \textbf{48.0}           &  65.6         & \textbf{55.1}           &  67.8          &  33.3         &  39.1         &  \textbf{56.3}         &  \textbf{39.6}         &  \textbf{80.8}       & 45.7         & 28.2          &  29.9        &  \textbf{31.8}      &  25.6        &  46.0        \\
\rowcolor{Green} HeuSCM (Ours)             & DeepLabv2   &   \textbf{92.7}
        &   \textbf{69.1}
          &    81.0
      & \textbf{43.9}          &   \textbf{34.0}
       &  42.8
         &   65.9
        &   48.8
       &  81.6
       &  \textbf{35.6}
      &  \textbf{93.3}
         &   47.3
       &   24.9
     &  79.4
      &   \textbf{55.5}
      &   \textbf{55.5}
       &   62.0
        &   14.4
      &  37.5
      &   \textbf{56.1}
      \\ 
\hline
DAFormer~\cite{hoyer2022daformer}             & DAFormer   & 71.4           & 50.0           & 79.1            & 42.1            & 33.1             & 55.5           & 40.0          & 50.0           & 72.2           & 35.4          & 68.2          & 54.3         & 17.5          & 83.0        & 71.2         & 74.8          & 80.9         & 41.2       & 31.5         & 55.3          \\ 
Refign~\cite{bruggemann2022refign}           & DAFormer   &  89.4          & 62.4            &  \textbf{85.5}           & 48.6            & 36.6             &  57.7          & 71.0           & 55.0           & 85.3           &  41.0          & 95.1         & 57.3         &  33.1         & 82.9        & 73.6          & 82.5          & 86.0         &  43.9      &  48.1        & 65.0         \\ 
VBLC~\cite{li2023vblc}             & DAFormer   &  88.5          &  57.6          &  81.9            & 41.2            &  35.2            &  58.0          &  \textbf{72.8}         & 57.5          &  71.7          &  39.3         &  82.1          &  62.2        &  36.2          &  \textbf{87.1}       &  82.6        & 86.6         &   84.1       &  41.6      &   44.9       &    63.7      \\
CoPT~\cite{mata2025copt}  & DAFormer     & 55.4   &  \textbf{71.7}         &  79.1         &   \textbf{57.7}          &   \textbf{47.2}          &    \textbf{62.9}          &   60.5         &  \textbf{65.2}         &    73.0    & 40.3     &   51.0          &  \textbf{68.9}        &   \textbf{46.6}        &   83.1        &    \textbf{82.7}        &    \textbf{90.8}     &    \textbf{88.2}      &   \textbf{47.8}       &      \textbf{58.6}   &     64.8                      \\
Instance-Warp~\cite{zheng2025instance}  & DAFormer   &  82.9         &  56.1          &  79.8          &  44.6           & 40.3             &  52.7          & 60.8          &  52.5          &  72.0          &  38.4         &  78.0         & 56.6         &  30.5         &   84.9      & 80.2         & 86.9          & 86.4         & 44.5       &  45.8        &  61.8         \\ 
\rowcolor{Green} HeuSCM (Ours)             & DAFormer    &  \textbf{90.8}         & 65.6           &  85.2        &  49.3          &   38.0        &   58.6        &   70.8        &  55.7        &  \textbf{85.4}       &  \textbf{42.4}      &  \textbf{95.1}        &  58.0        &  33.4     &  83.1       & 73.2       &  81.0        &   87.7        &  44.4       &  45.5       &   \textbf{65.4}      \\ 
\hline
HRDA~\cite{hoyer2022hrda}             & HRDA   & 86.5           & 52.5           & 83.7            & 50.6            & 34.8             & 61.4           & 72.7          & 60.7           & 72.3           & 39.7          & 81.9          & 65.5         & 45.5          & 88.1        & 84.7         & 82.8          & 74.3         & 48.4       & 53.7         & 65.2         \\
Refign~\cite{bruggemann2022refign}             & HRDA   &  94.7           &  76.5           &  87.1           &   52.9          &  43.4            &  62.1          & 77.4          &  65.8           & 86.4            &  44.0         & 95.3          &  64.5        & 42.4          &  87.9       &  85.5        &  90.4          & 89.8         &  47.7      &  56.3        &  71.1      \\
VBLC~\cite{li2023vblc}             & HRDA      & 89.5   & 61.8           & 84.9            &   46.3           & \textbf{48.5}            &  61.3             &  74.8          & 60.1          &  72.3          &  39.7          & 82.1          & 62.9          & 41.9          & 87.8           & 85.7        & 76.0         &  87.4         & 48.2          & 52.6       &  66.5                \\ 
CoDA~\cite{gong2024coda} & HRDA    & 93.3  & 72.6  & \textbf{88.3}  & \textbf{59.6}  & 48.2   & \textbf{63.3}  &  76.6   & 66.2    &  85.9   &  40.2  & 94.4   & \textbf{71.3}  & 48.5  & \textbf{90.5}  &  83.6 & 91.5 &  91.7   &  53.6   & 56.0  & 72.4        \\ 
ACSegFormer~\cite{liu2024domain} & HRDA     & \textbf{95.5}   & 77.1  &  87.5  & 56.9  & 45.8  & 59.5   &  79.4   &  66.2   & 86.3    &  42.9  & \textbf{95.3}   &  68.0 & 49.9  &  87.3  & \textbf{86.5}  & \textbf{94.1} &  91.7  & \textbf{55.3}  & 54.2  & 72.6        \\ 
CISS~\cite{sakaridis2025condition} & HRDA        & 92.3   & 69.9           &   85.2          &  50.5            &  41.3           &  60.7            &  76.0          &  61.2          &  73.1          &  41.2         &  82.4      &  67.5      &  45.0        &  89.6         &  83.8       &  88.3         &  89.1        &  53.2       &  54.5               &  68.7       \\ 
\rowcolor{Green} HeuSCM (Ours)             & HRDA    & 95.3   &   \textbf{77.9}         &  87.0           &  57.2            &  44.1          &     60.3         &  \textbf{79.2}          &  \textbf{66.7}        &    \textbf{86.5}       &   \textbf{44.7}       &  95.2     &   66.5     &  \textbf{50.7}        &   87.7      &  86.4     & 93.4        &  \textbf{91.8}      &  50.8      &  \textbf{60.9}         &  \textbf{72.7}  \\ \hline

\hline
\end{tabular}
}
\label{t:acdctval}
\end{table*}

\subsection{Categorical $\alpha$-Fairness for Policy Gradients}
\label{method2}

Due to inherent training bias, the model naturally favors certain semantic classes. However, traditional policy gradient calculation solely focuses on reward maximization, thereby neglecting the concept of fairness (\textit{i.e.}, the need for the agent to treat every semantic class equally). This issue is paramount in adverse weather autonomous driving scenarios, where robust and balanced segmentation performance across all classes is essential. 
To achieve fairness among the learning progress of different semantic classes, we adapt the concept of multi-agent fairness to our single-agent, multi-class problem.

We first define the value function for the $c$-th semantic class, $V_c^{\pi}(s)$, as the expected total discounted return that category $c$ can obtain, starting from state $s$ and following policy $\pi$:
\begin{equation}
  V_c^{\pi}(s) := \mathbb{E}_{\pi} \left[ \sum_{k=0}^{\infty} \gamma^k r_c(t+k) \mid S_t=s \right],
\end{equation}
where $r_c(t)$ is the reward signal computed specifically for class $c$ at time $t$. 

Inspired by the insight that models with enhanced transferability and discriminability yield superior target-domain performance, we propose a reward mechanism to explicitly quantify these two factors. Given the absence of target labels, we therefore compute an unsupervised, composite reward that synthesizes both, calculated as follows:
\begin{equation} 
\resizebox{0.9\columnwidth}{!}{$
\left\{
\begin{array}{l}
\begin{aligned}
& r_c(t) = \underbrace{ \frac{A_c^S(t) \cdot A_c^T(t)}{\|A_c^S(t)\| \cdot \|A_c^T(t)\|} }_{\text{Transferability}} + \lambda \cdot \underbrace{ \sum_{k \in \mathcal{C}, k \ne c} \left( 1 - \frac{A_c^T(t) \cdot A_k^T(t)}{\|A_c^T(t)\| \cdot \|A_k^T(t)\|} \right) }_{\text{Discriminability}}, \\
& A_c^S(t)=\frac{1}{\left|\Lambda_c^S(t)\right|} \sum_{(x_{s}, y_{s}) \in D_{S}} \mathbb{I}\{y_{s}=c\} \cdot f\left(x_{s}\right)|_c, \\
& A_c^T(t)=\frac{1}{\left|\Lambda_c^T(t)\right|} \sum_{(x_{t}, g_\theta(x_{t})) \in D_{T}} \mathbb{I}\{g_\theta(x_{t})=c\} \cdot f\left(x_{t}\right)|_c, \\
\end{aligned}
\end{array}
\right.
$}
\label{reward}
\end{equation}
where $|\Lambda_c^S(t)|$ and $|\Lambda_c^T(t)|$ is the number of pixels belongs to category $c$ in the source domain and target domain, respectively. $\mathbb{I}$ is an indicator function. $f(x)|_c$ denotes the feature output of our segmentation network $g_\theta$ for $c$.
$\lambda$ is a hyperparameter balancing the two objectives. 

To achieve fairness among the returns of all classes, we no longer optimize the standard policy objective $J^{\text{sum}}(\pi)$ (which sums the values of all classes and is thus susceptible to preference bias), but instead optimize a different global fairness objective 
$J_F(\pi)$:
\begin{equation}
\resizebox{0.9\columnwidth}{!}{$
\begin{aligned}
J_F(\pi) :=  & F(V_1^{\pi}(s), V_2^{\pi}(s), \dots, V_C^{\pi}(s))
 =  & \sum_{c=1}^{C} \frac{1}{1-\alpha} \left(V_c^{\pi}(s)\right)^{1-\alpha},
 \end{aligned}
 $}
\end{equation}
where $F$ is the function of each class's return, chosen to enforce fairness. 
To mitigate the model fairness issue exacerbated by the greedy nature of traditional RL reward mechanisms, the $\alpha$-fairness objective is employed as the objective function $F$.

When the $J_F(\pi)$ is applied to the policy gradient calculation, it corresponds to a weighted advantage function. The gradient of the policy $\pi_{\theta}$ is computed using a fairness-weighted aggregate advantage, $\tilde{A}_{\alpha}$ as follows:
\begin{equation}
\resizebox{0.88\columnwidth}{!}{$
\begin{aligned}
& \nabla_{\theta} J_F(\pi_{\theta})  = \mathbb{E}_{\pi} \left[ \nabla_{\theta} \log \pi_{\theta}(a_t \mid s_t) \cdot \tilde{A}_{\alpha}(s_t, a_t) \right] \\
& = \mathbb{E}_{\pi} \left[ \nabla_{\theta} \log \pi_{\theta}(a_t \mid s_t) \cdot \sum_{c=1}^{C} w_c(s_t) \cdot A_c(s_t, a_t) \right],
 \end{aligned}
 $}
\label{policy_gradient}
\end{equation}
where $\tilde{A}_{\alpha}(s_t, a_t)$ is our $\alpha$-fair advantage function, $A_c$ is the advantage function for semantic class $c$. And $w_c(s_t)$ denotes the fairness weight that is controlled by $\alpha$ and is inversely proportional to the $\alpha$-th power of the category's current value: $w_c(s_t) := V_c^{\pi}(s_t)^{-\alpha}$. By maximizing $J_F(\pi_{\theta})$, we can compel the policy $\pi$ to pursue a fairer and more balanced learning trajectory. The full training procedure is outlined in Algorithm 1.

\section{Experiments}

\subsection{Datasets and evaluation metrics}
The mean Intersection-over-Union (mIoU) is adopted as the evaluation metric, where a higher value indicates better performance. 
We validate the effectiveness of our method on 1) unsupervised domain adaptation semantic segmentation (UDA-SS) under adverse weather: Cityscapes$\rightarrow$ACDC domain adaptation; and 2) UDA night semantic segmentation: Cityscapes$\rightarrow$DarkZurich domain adaptation. Additionally, we evaluate the generalization ability of our method in the synthetic-to-real semantic segmentation, \textit{i.e.,} GTA5 $\rightarrow$ Cityscapes. 

\subsection{Experimental settings}
Our proposed framework is implemented using the PyTorch framework on an NVIDIA A800 GPU. 
DeepLab-v2~\cite{chen2017deeplab}, DAFormer~\cite{hoyer2022daformer}, and HRDA~\cite{hoyer2022hrda} as the backbone. 
$\lambda_{1}$ and $\lambda_{2}$ in Eq.~\ref{segloss} are both set to 1.0. And $\lambda$ in Eq.~\ref{reward} is set to 1.0.
Training is conducted for 60k iterations using 1024×1024 random crops from the Cityscapes and ACDC datasets. 
We train our model using the AdamW optimizer, setting the weight decay to 1e-4. For mixed image generation, we first employ our HeuSCM method, followed by standard augmentations including Color Jittering and Gaussian Blurring. Additionally, we adopt the rare class sampling strategy from~\cite{hoyer2022daformer} to mitigate the source domain's long-tail distribution, setting the $\alpha$ parameter to 0.999.

\subsection{Comparison with State-of-the-art Methods}

\subsubsection{Comparison on ACDC}
We present comparisons to several kinds of semantic segmentation methods, including 1) \textit{backbones}: DeepLab-v2~\cite{chen2017deeplab}, DAFormer~\cite{hoyer2022daformer}, and HRDA~\cite{hoyer2022hrda}; 
2) \textit{UDA-SS methods under Adverse Weather}: Refign~\cite{bruggemann2022refign}, CMA~\cite{bruggemann2023contrastive}, VBLC~\cite{li2023vblc}, CompUDA~\cite{zhengl2023compuda}, CoDA~\cite{gong2024coda} and ACSegFormer~\cite{liu2024domain}; and 3) \textit{general UDA-SS methods}: ATP~\cite{wang2024curriculum}, CISS~\cite{sakaridis2025condition}, Gaussian~\cite{yu2025contrastive}, CoPT~\cite{mata2025copt} and Instance-Warp~\cite{zheng2025instance}.
The quantitative results of mIoU performances on the ACDC test set and the ACDC val set are reported in Table~\ref{t:acdctest} and Table~\ref{t:acdctval}, respectively. We observe that our method consistently outperforms existing approaches across different backbones on both the ACDC test and validation sets. Notably, when built upon the HRDA backbone, our method achieves mIoU scores of 72.9 and 72.7 mIoU [\%] on the ACDC test and val sets, respectively, establishing state-of-the-art performance for Cityscapes $\rightarrow$ ACDC domain adaptation.

\begin{figure*}[t]
  \centering
   \includegraphics[width=1.0\textwidth]{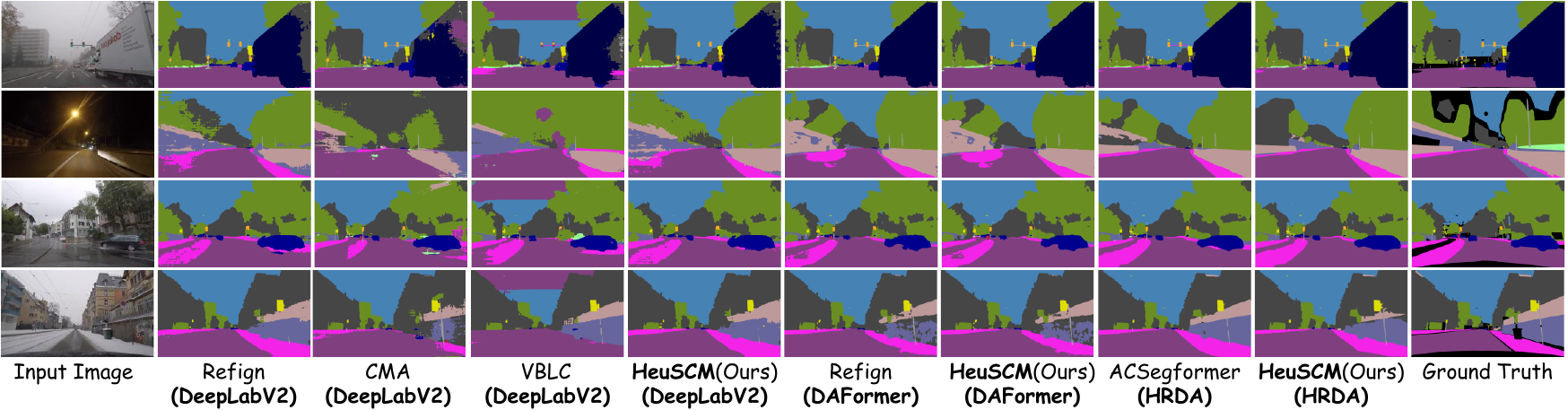}
   \caption{The qualitative comparison between our method and existing state-of-the-art methods based on DeepLabV2, DAFormer, and HRDA on the ACDC val. Compared with the existing state-of-the-art UDA method (\textit{e.g.,} Refign, CMA, VBLC, ACSegFormer), our method achieves better performance under the same backbone. Importantly, our method, based on HRDA, achieves the best performance, with predictions closely matching the ground truth, validating the effectiveness of our method.}
   \label{fig:experiments}
\end{figure*}
 
The qualitative comparison of our method with existing UDA on the ACDC val set is shown in Figure~\ref{fig:experiments}. Under the same backbone, our results are visually closer to the ground truth than existing UDA-SS under Adverse Weather methods. Furthermore, our HeuSCM, built upon the HRDA backbone, yields the best semantic segmentation performance, further validating its effectiveness.

\begin{figure}[t]
  \centering
   \includegraphics[width=0.5\textwidth]{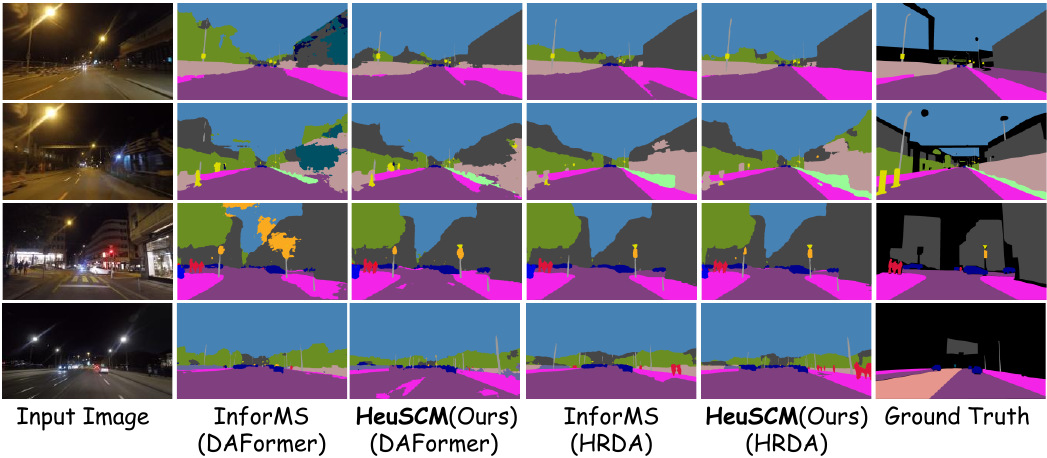}
   \caption{A qualitative comparison between our method and existing state-of-the-art approaches based on DAFormer and HRDA is conducted on the Dark Zurich val (Top two rows) and Nighttime Driving test (Bottom two rows). }
   \label{fig:experiments_dark}
\end{figure}

\subsubsection{Comparison on Dark Zurich}
To verify the effectiveness of our method on night scenes, we further conduct comparative experiments on the Dark Zurich-val dataset, and the results are shown in Table~\ref{t:night}. 
Notably, built upon HRDA, our method achieves state-of-the-art performance on Dark Zurich-val with 52.8 [\%] mIoU. The visualization results are presented in Figure~\ref{fig:experiments_dark}, which verify the effectiveness of our method.

\begin{table}
\caption{Comparison with state-of-the-art methods on the Dark Zurich-val set and Nighttime Driving test set.}
\resizebox{\linewidth}{!}{
\begin{tabular}{cccc}
\hline
\multirow{2}{*}{Method} & \multirow{2}{*}{Backbone}  & Dark Zurich-val & Nighttime Driving \\ \cline{3-4} 
                        &                            & mIoU            & mIoU          \\ \hline

DAFormer~\cite{hoyer2022daformer}            & DAFormer                    & 37.1               &  54.0          \\
InforMS~\cite{wang2023informative}      & DAFormer                   & 45.1          & 56.0    
\\ 
\rowcolor{Green} HeuSCM (Ours)         & DAFormer                  & \textbf{45.5}            & \textbf{56.7} 
\\ \hline
HRDA~\cite{hoyer2022hrda}            & HRDA                    &  42.1              & 54.1          \\
InforMS~\cite{wang2023informative}         & HRDA               & 52.5            & 58.5  \\
\rowcolor{Green} HeuSCM (Ours)         & HRDA                   & \textbf{52.8}            & \textbf{59.3} 
\\ \hline
\end{tabular}
}
\label{t:night}
\end{table}

\begin{table}[]
\caption{Ablation Study on several model variants of our method on the ACDC val. LSRL, SKFEN, C$\alpha$PG are Low-dimensional State Representation Learning in~\ref{low-dimensional}, Semantic Key Feature Extraction Network in~\ref{SKFEN}, Categorical $\alpha$-Fairness for Policy Gradients in~\ref{method2}, respectively.}
\renewcommand\arraystretch{1.1}
\centering
  \resizebox{\linewidth}{!}{
\begin{tabular}{lcccccccc}
\hline
  LSRL        &                                 & \checkmark &                           &                           & \checkmark & \checkmark &                           & \checkmark \\
                            SKFEN       &                                 &                           & \checkmark &                           & \checkmark &                           & \checkmark & \checkmark \\
                            C$\alpha$PG &                                 &                           &                           & \checkmark &                           & \checkmark & \checkmark & \checkmark \\ \hline
                           mIoU       & \multicolumn{1}{c}{71.1 (+0.0)} & 72.2 (+1.1)               &        71.7 (+0.6)                   &  71.6 (+0.5)                         & 72.3 (+1.2)               &     72.0 (+0.9)                      &  72.2 (+1.1)                         & \textbf{72.7 (+1.6)}               \\ \hline
\end{tabular}
 }
\label{ablation_experiments}
\end{table}

\begin{table}[ht]
\caption{Comparison of semantic segmentation performance on GTA5 $\rightarrow$ Cityscapes on the Cityscapes validation set. 
Best results are in \textbf{bold}.}
\label{t:gta2city}
\centering
\resizebox{\linewidth}{!}{
\begin{tabular}{ccc}
\hline
\multirow{2}{*}{Method} & \multirow{2}{*}{Backbone} & GTA5 $\rightarrow$ Cityscapes \\ \cline{3-3} 
 &   & \textbf{mIoU} \\
\hline
IAST~\cite{mei2020instance} & Deeplab-v2 & 52.2  \\
\rowcolor{Green} IAST + HCSP (Ours) & Deeplab-v2 & 52.4 \\ \hline
HIAST~\cite{zhu2025hard} & Deeplab-v2  & 56.3  \\
\rowcolor{Green} HIAST + HCSP (Ours) & Deeplab-v2  & \textbf{56.5} \\ \hline
\end{tabular}
}
\label{t:syclear}
\end{table}

\subsubsection{Comparison on Nighttime Driving}
To show our method's generalization on night scenes, we also evaluate our approach on the Nighttime Driving test set in Table~\ref{t:night} when performing cityscapes$\rightarrow$DarkZurich domain adaptation, with sample visualization results presented in Figure~\ref{fig:experiments_dark}. With HRDA as the backbone, our method consistently achieves the highest performance on this dataset, reaching 59.3 [\%] mIoU. These results confirm the strong generalization capability of our method on the Nighttime Driving dataset.  

\subsection{Ablation Study}
In this section, we validate the effectiveness of our three core innovations. For the Cityscapes $\to$ ACDC domain adaptation, we trained several model variants of our HeuSCM (HRDA) and evaluated their performance on the ACDC val set, as shown in Table~\ref{ablation_experiments}. We adopt Refign \cite{bruggemann2022refign} as our baseline and incrementally incorporate our proposed components. The results show that adding LSRL, SKFEN, and C$\alpha$PG individually yields performance gains of 1.1, 0.6, and 0.5 mIoU, respectively. Furthermore, combining two components (LSRL and SKFEN, or LSRL and C$\alpha$PG, or SKFEN and C$\alpha$PG) leads to improvements of 1.2, 0.9, and 1.1 mIoU. And full integration of all designs achieves a top performance of 72.7 [\%] mIoU, confirming the collective effectiveness and indispensability of each module.

\subsection{Generalization Study}
To verify the generalization of our proposed Heuristic Class Sampling Policy (HCSP), we replace the sampling strategy of existing hard-class mining methods with our HCSP for synthetic-to-real semantic segmentation. Experimental results are summarized in Table~\ref{t:syclear}. Notably, our HCSP consistently yields significant performance gains, confirming its superior generalization capability.
\section{Conclusion}
In this paper, we introduce HeuSCM, a novel reinforcement learning framework for unsupervised domain adaptation semantic segmentation. Different from existing class curriculum learning that relies on predefined, human-designed heuristics, we proposed a paradigm shift from ``designing" curricula to ``learning" them. Our framework employs an autonomous agent guided by two key technical innovations: (1) a High-dimensional Semantic State Extraction that perceives the learning status of the semantic segmentation model, and (2) a Categorical $\alpha$-Fairness for Policy Gradients that achieves equitable rewards across semantic classes.
Extensive experiments demonstrated that our method achieves state-of-the-art performance on challenging adverse weather segmentation benchmarks. Furthermore, its superior results on synthetic-to-real semantic segmentation validate the strong generalization capability of our approach. 
We believe our work opens a promising new avenue for unsupervised domain adaptation, demonstrating that learned, strategic, multi-objective scheduling policies can significantly outperform traditional fixed-heuristic methods in complex adaptation scenarios.

\section*{Acknowledgement}
This work was funded by the National Natural Science Foundation of China (Grant No. 62571379) and the Hubei Provincial Key Research and Development Program (Grant No. 2024BAB050). The numerical calculations in this paper have been done on the supercomputing system in the Supercomputing Center of Wuhan University.

{
    \small
    \bibliographystyle{ieeenat_fullname}
    \bibliography{main}
}

\end{document}